\documentclass{article}

\usepackage[preprint]{neurips_2025}

\usepackage[utf8]{inputenc}
\usepackage[T1]{fontenc}
\usepackage{hyperref}
\usepackage{url}
\usepackage{booktabs}
\usepackage{amsmath,amssymb,amsfonts}
\usepackage{nicefrac}
\usepackage{microtype}
\usepackage{xcolor}
\usepackage{graphicx}
\usepackage{lipsum} 

\usepackage{longtable}
\usepackage{booktabs}
\usepackage{array}

\usepackage{caption}

\title{Ultrasound Vision–Language Alignment via Contrastive Learning}

\author{
  Zhuoyang Lyu \thanks{BMI 702: AI in Medicine II (AIM2).} \thanks{Equal contribution} \\
  Department of Biomedical Informatics \\
  Harvard Medical School \\
  Boston, MA, USA \\
  \texttt{zhuoyang\_lyu@hms.harvard.edu}
  \And
  Ruirui Lan \footnotemark[1] \footnotemark[2] \\
  Department of Biomedical Informatics \\
  Harvard Medical School \\
  Boston, MA, USA \\
  \texttt{ruirui\_lan@hms.harvard.edu}
  \And
  Tongxin Wang \footnotemark[1] \footnotemark[2] \\
  Department of Biomedical Informatics \\
  Harvard Medical School \\
  Boston, MA, USA \\
  \texttt{tongxin\_wang@hms.harvard.edu}
  \And
  Yiyang Zhang \footnotemark[1] \footnotemark[2]\\
  Department of Biostatistics \\
  Harvard T. H. Chan School of Public Health \\
  Boston, MA, USA \\
  \texttt{yiyangzhang@hsph.harvard.edu}
}

\begin{document}

\maketitle


\begin{abstract}
Ultrasound foundation models have achieved strong performance on structured prediction tasks but remain exclusively vision-based, limiting zero-shot and few-shot transfer to novel tasks where task-specific annotation is scarce. We address this gap with EchoCare-CLIP, a CLIP-style dual-encoder contrastive framework that aligns ultrasound images with clinical text in a
shared embedding space. We curate a multi-organ corpus of over 16K image-text pairs spanning breast, liver, lung, and thyroid, with over 78\% of captions derived from expert-annotated reports, and complement the remainder with a three-tier template-based and LLM-based caption generation pipeline. We evaluate model configurations spanning two text encoder
families (CLIP\cite{radford2021clip, clipvitbasepatch32_hf}, BioClinicalBERT\cite{alsentzer2019bioclinicalbert, alsentzer2019bioclinicalbert_hf}) and two caption strategies (template-based, LLM-generated) against OpenAI CLIP\cite{radford2021clip} and BiomedCLIP\cite{zhang2025biomedclipmultimodalbiomedicalfoundation} baselines. Our trained models consistently improve cross-modal alignment over baselines, with the best configuration achieving a paired alignment score of 0.682. However, stronger alignment does not guarantee better downstream performance: CLIP-based variants with partial fine-tuning achieve the strongest zero-shot classification on external held-out datasets (0.709 on BUSI\cite{al-dhabyani2020busi}; 0.626 on AULI\cite{xu2022auli}), while full end-to-end fine-tuning degrades transfer due to overfitting. On linear probing and few-shot adaptation, model rankings are dataset-dependent, reflecting a trade-off between domain adaptation and representational generalizability. We further show that template-based captions match or outperform LLM-generated captions, suggesting lexical diversity is not a proxy for caption quality for contrastive ultrasound pretraining. Taken together, our results demonstrate that ultrasound vision-language alignment is achievable from public data alone, but robust clinical transfer requires careful balancing of domain adaptation, encoder capacity, and caption supervision quality. \footnote{Code available at \url{https://github.com/TongxinWang0405/BMI702_project.git}}
\end{abstract}

\section{Introduction}
\label{sec:introduction}

Ultrasound is the most widely used medical imaging modality worldwide\cite{wells2006ultrasound, jiang2023ultrasound_med_background}. Yet its 
interpretation remains highly operator-dependent and difficult to standardize, making expert-annotated data expensive and difficult to collect at scale \cite{jiang2023ultrasound_med_background, Strauss2007inter_intra_variability_sonograph, Itani2019inter_variability_sonograph}. Recent ultrasound foundation models -- EchoCare~\cite{zhang2025echocare} and USFM~\cite{jiao2024usfm} -- have demonstrated strong performance on structured prediction tasks through large-scale visual pretraining, but rely exclusively on image supervision and lack text alignment. This constrains adaptation to settings with sufficient task-specific labeled data, limiting practical utility where annotation budgets are tightest\cite{jiang2023ultrasound_med_background, litjens2017survey_med_image}. We ask whether contrastive vision-language pretraining can close this gap, enabling zero-shot classification and transfer to unseen datasets without any task-specific annotation.

Contrastive vision-language pretraining has proven effective at bridging this gap in other medical domains. CLIP~\cite{radford2021clip} showed that image-text contrastive training on large-scale pairs yields strong zero-shot transfer; ConVIRT~\cite{Zhang2020convirt} and CONCH~\cite{lu2024conch} extended this paradigm to radiology and pathology, respectively. No analogous model exists for ultrasound, largely because paired image-report datasets are exceptionally scarce in this modality.

We address this gap with \textit{EchoCare-CLIP}, a CLIP-style dual-encoder 
contrastive framework that aligns ultrasound images with clinical text in a shared 
embedding space. To construct paired training data, we aggregate a multi-organ 
corpus of publicly available ultrasound images. We systematically evaluate four training configurations across two text encoder families (CLIP\cite{radford2021clip, clipvitbasepatch32_hf}, BioClinicalBERT\cite{alsentzer2019bioclinicalbert, alsentzer2019bioclinicalbert_hf}) and two caption strategies (template-based, LLM-generated), benchmarking against OpenAI CLIP\cite{radford2021clip, clipvitbasepatch32_hf} and BiomedCLIP~\cite{zhang2025biomedclipmultimodalbiomedicalfoundation, biomedclip_hf} baselines.

\noindent Our main contributions are:
\begin{itemize}
    \item We curate a multi-organ ultrasound corpus of over 16K image-text 
    pairs spanning breast, liver, lung, and thyroid, with over 78\% of captions 
    derived from expert-annotated reports.
    \item We propose EchoCare-CLIP and compare four fine-tuning configurations, 
    demonstrating improvements in cross-modal alignment and zero-shot 
    classification over general-domain and biomedical baselines.
    \item We introduced a three-tier template based and a LLM-based caption generation pipeline and show that linguistic diversity alone is not the dominant factor in zero-shot transfer performance.
    \item We provide systematic evaluation of linear probing, few-shot 
    adaptation (1\%, 5\%, 10\% label fractions), and cross-modal retrieval for 
    our ultrasound vision-language model.
\end{itemize}

\section{Background}
\label{sec:background}

\paragraph{Vision-language pretraining.}
CLIP~\cite{radford2021clip} demonstrated that contrastive training on massive datasets of internet image-text pairs yields strong zero-shot transfer through cosine similarity in a shared embedding space. Subsequent work extended this paradigm to the medical domain: ConVIRT~\cite{Zhang2020convirt} applied bidirectional contrastive learning to chest X-ray-report pairs, and BiomedCLIP~\cite{zhang2025biomedclipmultimodalbiomedicalfoundation} scaled biomedical vision-language pretraining to millions of image-text pairs drawn from PubMed Central. CONCH~\cite{lu2024conch} further demonstrated that domain-specific image-text alignment substantially improves downstream performance in computational pathology over general-domain baselines. The consistent finding across these works is that pretraining domain matters: models trained on data closer to the target domain generalize better. Our work follows this paradigm for ultrasound, a domain absent from all existing vision-language pretraining corpora at meaningful scale.

\paragraph{Ultrasound foundation models.}
EchoCare~\cite{zhang2025echocare} pretrains a SwinTransformerV2 on approximately 4.5M ultrasound images, showing that ultrasound-domain visual pretraining outperforms transfer from natural image models on structured prediction tasks. USFM~\cite{jiao2024usfm} similarly leverages large-scale ultrasound pretraining to improve cross-organ robustness. Both models follow a vision-only paradigm with no text supervision, constraining adaptation to settings with sufficient labeled data. We build directly on EchoCare's visual backbone and introduce text alignment as a complementary pretraining signal.

\paragraph{Data scarcity in ultrasound.}
A key barrier to ultrasound vision-language research is the near-absence of publicly available paired image-report datasets\cite{wells2006ultrasound, jiang2023ultrasound_med_background}. Unlike chest X-ray (MIMIC-CXR~\cite{johnson2019mimiccxr}) or CT, ultrasound studies rarely produce free-text reports electronically linked to individual frames\cite{wells2006ultrasound, jiang2023ultrasound_med_background}. This scarcity directly explains the limited development of ultrasound VLMs relative to other imaging modalities, and motivates our caption generation approach as a practical surrogate for paired clinical supervision.

\section{Methods}
\label{sec:methods}

\subsection{Dataset}

Our internal training/evaluation corpus comprises 16,464 ultrasound image--text pairs from six publicly available datasets spanning breast, lung, thyroid, and liver. We additionally reserve BUSI and AULI as external held-out datasets that are not used for training, validation, or model selection. We included samples when the image contained usable ultrasound content and had either expert annotation, report-style text, structured metadata, or diagnostic labels sufficient to construct an image-text pair. Details in Table ~\ref{tab:datasets}.

Six datasets were used for internal training, validation, and internal held-out testing: BUS/BUS-Lesion~\cite{yu2026buscot} with 5,163 breast image-report pairs, BrEaST~\cite{pawlowska2024breast} with 248 breast ultrasound samples, Lung US~\cite{katumba2025lung} with 1,062 lung ultrasound samples, AUITD~\cite{azouz2023auitd} with 2,289 thyroid ultrasound samples, Report Gen-Thyroid~\cite{li2024usreport} with 4,914 thyroid ultrasound samples, and Report Gen-Liver~\cite{li2024usreport} with 2,788 liver ultrasound samples.The six internal datasets were used for training, validation, and internal held-out testing with stratified sampling by dataset source. Two datasets were reserved as external held-out evaluation sets and were not used for training or validation: BUSI~\cite{al-dhabyani2020busi} with 780 breast ultrasound images and AULI~\cite{xu2022auli} with 735 liver ultrasound images. This setup was designed to evaluate both in-distribution alignment and external transfer on independently sourced breast and liver datasets.

Caption sources differed across datasets. BUS/BUS-Lesion and the Report-Gen thyroid and liver datasets include expert annotations or report-style text, which together account for more than 78\% of the internal training corpus. For datasets without paired free-text reports, image--text pairs were constructed by pairing each processed ultrasound image with a generated caption derived from available metadata or diagnostic labels (~\ref{subsec:caption_generation}). Downstream labels were defined from the available dataset annotations: organ labels were assigned by dataset source, breast/thyroid/liver condition labels were mapped to normal, benign, or malignant when available, and Lung US labels were mapped to healthy, Covid, or other disease. Template-based captions were used as the primary caption format for the main model comparisons, while LLM-generated captions were evaluated separately as an ablation.

Preprocessing followed a standardized pipeline across datasets. Images were deduplicated, blank or low-information frames were removed using an intensity threshold, videos were temporally subsampled when applicable, and all images were resized to 256$\times$256 pixels and normalized using training-set statistics. Text preprocessing included de-identification, sentence segmentation, abbreviation normalization, and caption generation only when expert annotation was unavailable.

\begin{table}[ht]
  \centering
  \caption{Summary of ultrasound image-text datasets used for model training and evaluation.}
  \label{tab:datasets}
  \renewcommand{\arraystretch}{1.25}
  \setlength{\tabcolsep}{3pt}
  \begin{tabular}{
      l                          
      r                          
      r                          
      l                          
      l                          
      l                          
    }
    \toprule
    \textbf{Dataset} &
    \textbf{\# Pairs} &
    \textbf{\% Internal} &
    \textbf{Label Type} &
    \textbf{Caption Type} &
    \textbf{Tissue} \\
    
    \midrule
    \multicolumn{6}{l}{%
      \small\itshape Internal datasets used for training} \\
    \midrule

    BUS
      & 5{,}163 & 31.4\%
      & N/A
      & Expert annotation
      & Breast \\

    BrEaST
      & 248     &  1.5\%
      & Benign/Malignant
      & Metadata
      & Breast \\

    Lung US
      & 1{,}062 &  6.6\%
      & Healthy/Covid/Other
      & Metadata
      & Lung \\

    AUITD
      & 2{,}289 & 13.9\%
      & Normal/Benign/Malignant
      & Tissue\,+\,Cond.
      & Thyroid \\

    Rep. Gen-Thyroid
      & 4{,}914 & 29.8\%
      & N/A
      & Expert annotation
      & Thyroid \\

    Rep. Gen-Liver
      & 2{,}788 & 16.9\%
      & N/A
      & Expert annotation
      & Liver \\

    \midrule
    \multicolumn{6}{l}{%
      \small\itshape Held-out external datasets (not used in training)} \\
    \midrule

    AULI
      & 735     & ---
      & Normal/Benign/Malignant
      & Tissue\,+\,Cond.
      & Liver \\

    BUSI
      & 780     & ---
      & Normal/Benign/Malignant
      & Tissue\,+\,Cond.
      & Breast \\

    \bottomrule
  \end{tabular}
\end{table}

\subsection{Baseline Models}

To evaluate the effectiveness of our ultrasound-specific contrastive pretraining, we benchmark our EchoCare-CLIP configurations against two established off-the-shelf vision-language foundation models:

\paragraph{OpenAI CLIP} 
We evaluate the standard \texttt{ViT-B/32} OpenAI CLIP model~\cite{radford2021clip, clipvitbasepatch32_hf}. This model was pretrained via contrastive learning on WIT-400M, a massive corpus of 400 million general-domain image-text pairs collected from the internet. We include OpenAI CLIP to serve as our general-domain baseline, representing the zero-shot capabilities of large-scale vision-language alignment without any medical or ultrasound-specific adaptation.

\paragraph{BiomedCLIP.} 
We also evaluate BiomedCLIP~\cite{zhang2025biomedclipmultimodalbiomedicalfoundation, biomedclip_hf}, a domain-specific model pretrained on PMC-15M—a dataset of 15 million biomedical image-text pairs extracted from PubMed Central articles. While this model is highly optimized for the broader medical domain (heavily featuring radiology, pathology, and microscopy), its pretraining corpus contains minimal ultrasound data. BiomedCLIP serves as an in-domain (medical) but out-of-modality (non-ultrasound) baseline, allowing us to isolate the performance gains specifically attributable to ultrasound-targeted cross-modal alignment.

\subsection{EchoCare-CLIP Model}

We propose a CLIP-style bidirectional contrastive learning framework for ultrasound vision-language alignment. A dual-encoder model maps paired ultrasound images and clinical text descriptions into a shared 256-dimensional embedding space, enabling zero-shot and few-shot transfer to downstream tasks without task-specific fine-tuning.

The image encoder uses the EchoCare SwinTransformerV2 backbone~\cite{zhang2025echocare, liu2022swinv2} (${\sim}120$M parameters) pretrained on approximately 4.5M ultrasound images, taking $256{\times}256$ input and producing a 2048-dimensional global-average-pooled representation. The text encoder uses the pretrained \texttt{openai/clip-vit-base-patch32} CLIP text model~\cite{radford2021clip, clipvitbasepatch32_hf} (${\sim}65$M parameters), encoding captions of up to 77 tokens into a 512-dimensional representation via the \texttt{[EOS]} token pooled output. As an alternative text encoder, we also evaluated \texttt{BioClinicalBERT}~\cite{alsentzer2019bioclinicalbert, alsentzer2019bioclinicalbert_hf} (${\sim}110$M parameters), a BERT-based model pretrained on clinical notes from the MIMIC-III database, which encodes captions into a 768-dimensional representation via the \texttt{[CLS]} token pooled output. Both encoders are followed by a two-layer MLP projection head (Linear $\to$ ReLU $\to$ Linear, ${\sim}5$M parameters) that maps into the shared 256-dimensional latent space, with L2 normalization applied to all output embeddings.

\subsection{Model Training}
All models are trained with the symmetric InfoNCE~\cite{oord2018infonce} contrastive loss:
\begin{equation}
    \mathcal{L} = \frac{1}{2}\left(\mathcal{L}_{\text{i}{\to}\text{t}} + \mathcal{L}_{\text{t}{\to}\text{i}}\right),
\end{equation}
where $\mathcal{L}_{\text{i}{\to}\text{t}}$ and $\mathcal{L}_{\text{t}{\to}\text{i}}$ are cross-entropy losses over the similarity logit matrix for image-to-text and text-to-image directions, respectively. The temperature parameter $\tau$ is initialized to $0.07$ and learned as $\exp(\log\tau)$ during training.

Input images are resized to $256{\times}256$ pixels and normalized using training set statistics computed from the aggregated corpus. During training, random crop (from $288{\times}288$) and random horizontal flip are applied as data augmentation; validation and test images use deterministic center crop only. Text captions are tokenized using the tokenizer corresponding to the text encoder: captions are tokenized with the CLIP tokenizer, truncated or padded to a maximum sequence length of 77 tokens, when using the CLIP text encoder; when using \texttt{Bio\_ClinicalBERT}, captions are tokenized with the corresponding \texttt{BioClinicalBERT} tokenizer, truncated or padded to a maximum sequence length of 128 tokens.

All experiments use AdamW optimizer~\cite{loshchilov2019adamw} ($\text{lr}=1{\times}10^{-4}$, weight decay $=1{\times}10^{-4}$) with cosine annealing learning rate scheduling~\cite{loshchilov2017sgdr} over 20 epochs and a batch size of 32.

\subsection{Caption Generation}
\label{subsec:caption_generation}

\subsubsection{Template-Based Captions}

Because few public ultrasound datasets provide free-text clinical reports paired with images, we developed a three-tier template caption generation pipeline to construct synthetic paired text supervision from available structured metadata and classification labels. Each tier progressively incorporates richer clinical context:

\begin{description}
    \item[Tier 1 — Tissue label only.]
        Encodes the anatomical region and coarse diagnostic category.
        \textit{Example:} ``An ultrasound image of breast with benign findings.''

    \item[Tier 2 — Tissue and condition labels.]
        Adds imaging view and laterality where available.
        \textit{Example:} ``This ultrasound demonstrates upper lateral longitudinal view of the left lung exhibiting features of probable Covid.''

    \item[Tier 3 — metadata.]
        Incorporates patient demographics, imaging view, and structured findings.
        \textit{Example:} ``A 50-year-old male patient. Ultrasound of lower lateral transverse view of the left lung reveals >3 B-lines findings, consistent with probably Covid.''
\end{description}

The complete set of 30 templates, organised by tier and anatomical domain, is
provided in the Extended Materials section ~\ref{subsec:caption_templates}.

\subsubsection{LLM-Based Caption Generation}

To increase lexical diversity beyond the fixed vocabulary of structural templates, we prompted a large language model to rewrite each template caption into more naturalistic clinical prose. We used \textsc{Qwen3-4B-Instruct-2507}\cite{Qwen3-4B-Instruct-2507}. For each source caption, the model generated
three distinct rewrites; one caption was randomly sampled from the three to maximize corpus-level diversity. Details in Extended Materials Section ~\ref{subsec:llm_based_caption_generation}

\subsection{Experiments and Evaluations} 
To isolate the contribution of each encoder component, we compare four training conditions (Table~\ref{tab:model_architecture}): (1) projection heads only, (2) projection heads + image encoder, 
(3) projection heads + text encoder, and (4) projection heads + both encoders.

The dataset was partitioned into train (70\%), validation (15\%), and test (15\%) splits using stratified sampling by dataset source to ensure proportional representation across all splits. Model selection was performed on the validation set; final results are reported on the internal held-out test split as well as external testing datasets.

We evaluate model performance using four complementary metrics suited to different aspects of the task. For zero-shot classification, we report \textbf{Accuracy}, computed as the fraction of correctly classified samples when the predicted class is determined by the highest cosine similarity to class-name text prompts. For linear probing and few-shot adaptation experiments, we report \textbf{AUROC} (Area Under the Receiver Operating Characteristic Curve), which is threshold-independent and better captures discriminative performance under class imbalance. For cross-modal retrieval, we report \textbf{Recall@K} (R@1, R@5, R@10) in both image-to-text (I2T) and text-to-image (T2I) directions, measuring the fraction of queries for which the correct match appears within the top-$K$ retrieved items (see Extended Materials Section ~\ref{subsec:retrieval}). Finally, we report the \textbf{alignment score} as a secondary diagnostic metric to track the quality of image-text correspondence in the shared embedding space after training. It is defined as the mean paired cosine similarity between L2-normalized image and text embeddings:

\begin{equation}
    \text{Alignment Score} = \frac{1}{N}\sum_{i=1}^{N} \frac{\mathbf{z}_i^{\text{img}} \cdot \mathbf{z}_i^{\text{txt}}}{\|\mathbf{z}_i^{\text{img}}\|\,\|\mathbf{z}_i^{\text{txt}}\|},
\end{equation}

\noindent where $\mathbf{z}_i^{\text{img}}$ and $\mathbf{z}_i^{\text{txt}}$ are the projected embeddings for the $i$-th matched image-text pair. Since both embeddings are L2-normalized, this reduces to the dot product of the unit vectors, with the score ranging in $[-1, 1]$ where higher values indicate tighter alignment.

All experiments were conducted on a single NVIDIA A100 GPU using PyTorch, with model loading and tokenization handled via the Hugging Face Transformers library.


\section{Results}
\label{sec:results}

We evaluate a total of ten model architectures, including two off-the-shelf benchmark models and eight task-adapted variants with different encoder configurations. 
Table~\ref{tab:model_architecture} summarizes the architectural design of all models, including the choice of image and text encoders, the components that were fine-tuned, and the number of trainable parameters.

In the following sections, we compare model performance across multiple evaluation settings, including zero-shot classification, linear probing, and few-shot adaptation. Results for cross-modal retrieval are in Extended Materials Section~\ref{subsec:retrieval}.

\begin{table}[t]
\centering
\small
\setlength{\tabcolsep}{4pt}
\renewcommand{\arraystretch}{1.15}
\caption{Model architectures and training configurations.}
\label{tab:model_architecture}
\begin{tabular}{llll}
\toprule
\textbf{Model} & \textbf{Encoder} & \textbf{Trained Components} & \textbf{Params} \\
\midrule
\multicolumn{4}{l}{\textit{Benchmark models}} \\
OpenAI CLIP & ViT-B/32 + CLIP Text & None & None \\
BiomedCLIP & ViT-B/16 + PubMedBERT & None & None \\
\midrule
\multicolumn{4}{l}{\textit{CLIP text encoder variants}} \\
MLP heads only & EchoCare + CLIP Text & Projection heads & $\sim$5M \\
MLP + image encoder & EchoCare + CLIP Text & Image encoder + heads & $\sim$101M \\
MLP + text encoder & EchoCare + CLIP Text & Text encoder + heads & $\sim$69M \\
MLP + image + text encoder & EchoCare + CLIP Text & Both encoders + heads & $\sim$165M \\
\midrule
\multicolumn{4}{l}{\textit{BioClinicalBERT text encoder variants}} \\
MLP heads only (BERT) & EchoCare + BioClinicalBERT & Projection heads & $\sim$6M \\
MLP + image encoder (BERT) & EchoCare + BioClinicalBERT & Image encoder + heads & $\sim$102M \\
MLP + text encoder (BERT) & EchoCare + BioClinicalBERT & Text encoder + heads & $\sim$114M \\
MLP + image + text encoder (BERT) & EchoCare + BioClinicalBERT & Both encoders + heads & $\sim$210M \\
\bottomrule
\end{tabular}
\end{table}

\subsection{Internal Test Split}
\begin{table}[t]
\centering
\small
\setlength{\tabcolsep}{5pt}
\renewcommand{\arraystretch}{1.15}
\caption{Model performance on held-out test set.}
\label{tab:model_performance}
\begin{tabular}{lccc}
\toprule
\textbf{Model} & \textbf{Paired Align.} & \textbf{Cross Align.} & \textbf{Cls. Acc.} \\
\midrule
OpenAI CLIP & 0.287 & 0.285 & 0.040 \\
BiomedCLIP & 0.382 & 0.319 & 0.414 \\
\midrule
MLP heads only & 0.323 & -0.007 & 0.336 \\
MLP + image encoder & 0.364 & -0.021 & 0.268 \\
MLP + text encoder & 0.465 & 0.026 & 0.264 \\
MLP + image + text encoder & 0.534 & 0.151 & 0.091 \\
\midrule
MLP heads only (BERT) & 0.484 & 0.157 & 0.447 \\
MLP + image encoder (BERT) & 0.592 & 0.170 & 0.492 \\
MLP + text encoder (BERT) & 0.501 & 0.027 & 0.335 \\
MLP + image + text encoder (BERT) & \textbf{0.682} & \textbf{0.014} & \textbf{0.495} \\
\bottomrule
\end{tabular}
\end{table}
We first evaluate all models on the held-out test split using two complementary perspectives: 
(1) cross-modal alignment quality, and 
(2) downstream zero-shot classification performance. 
Table~\ref{tab:model_performance} summarizes the main quantitative results across all models.

Among our trained models, alignment performance improves consistently as more components are unfreezed. This likely reflects more trainable parameters allow the models to fit to train data more accurately based on alignment score.
In particular, jointly training both image and text encoders leads to the highest paired alignment score for each text encoder variants 
(e.g., 0.534 for CLIP-based models and 0.682 for BERT-based models). 
This trend shows that full cross-modal co-adaptation is critical for maximizing embedding similarity on paired data.

However, this improvement does not translate directly into classification performance. 
For example, the CLIP-based model with both encoders trained achieves the highest alignment score (0.534) 
but suffers a sharp drop in classification accuracy (0.091). 
This mismatch suggests that the model is overfitting to the alignment objective since it learns to maximize similarity 
for training-style captions without preserving critical discriminative structure between clinical classes.

In contrast, models using BioClinicalBERT as the text encoder demonstrate a more favorable balance between alignment and classification. 
The best-performing model in this family achieves the highest alignment score overall (0.682) 
while also maintaining the best classification accuracy (0.495). 
As a result, according to the internal held-out test split data, BERT-based models yield higher classification accuracy compared to CLIP-text variants,
indicating that domain-specific language representations better capture clinically relevant semantics.

\subsubsection{Per Organ Result}

\begin{figure}[t]
    \centering
    \includegraphics[width=1\linewidth]{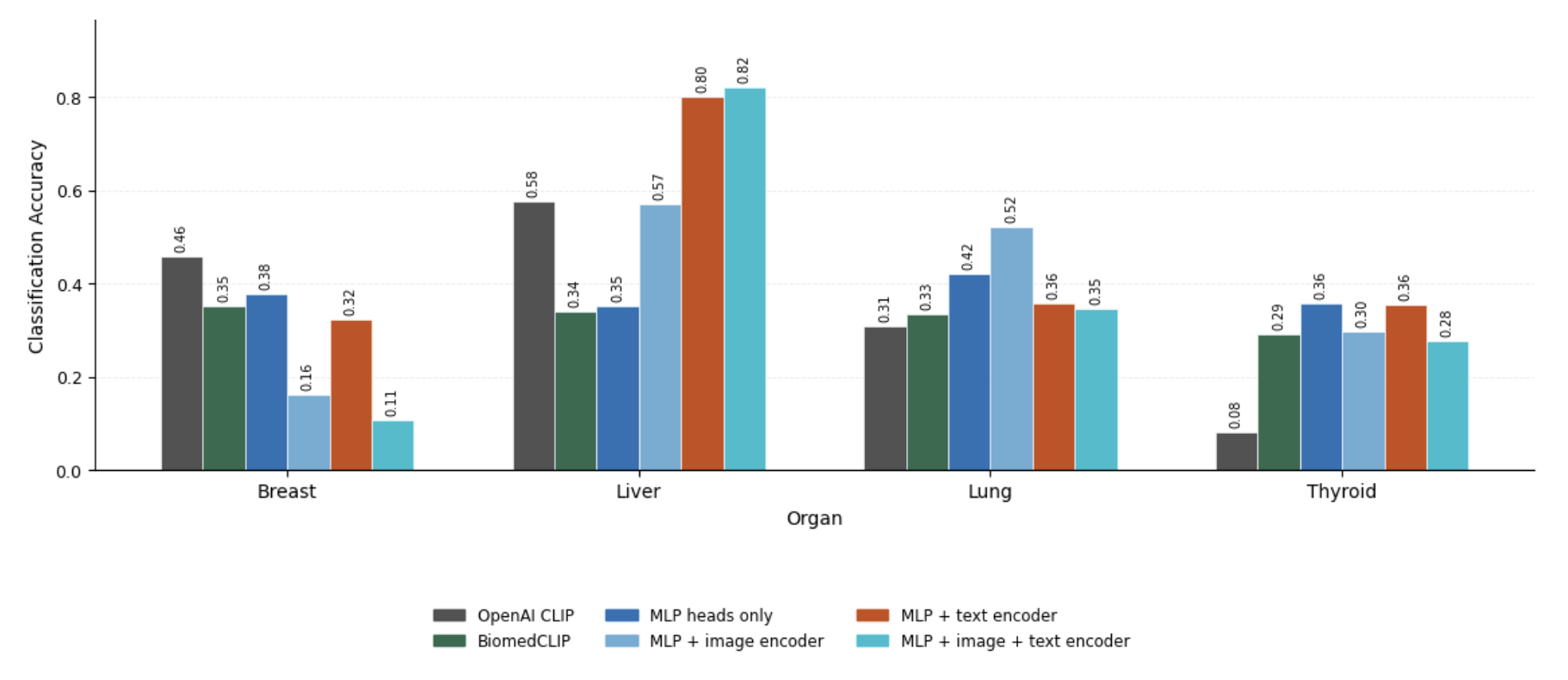}
    \includegraphics[width=1\linewidth]{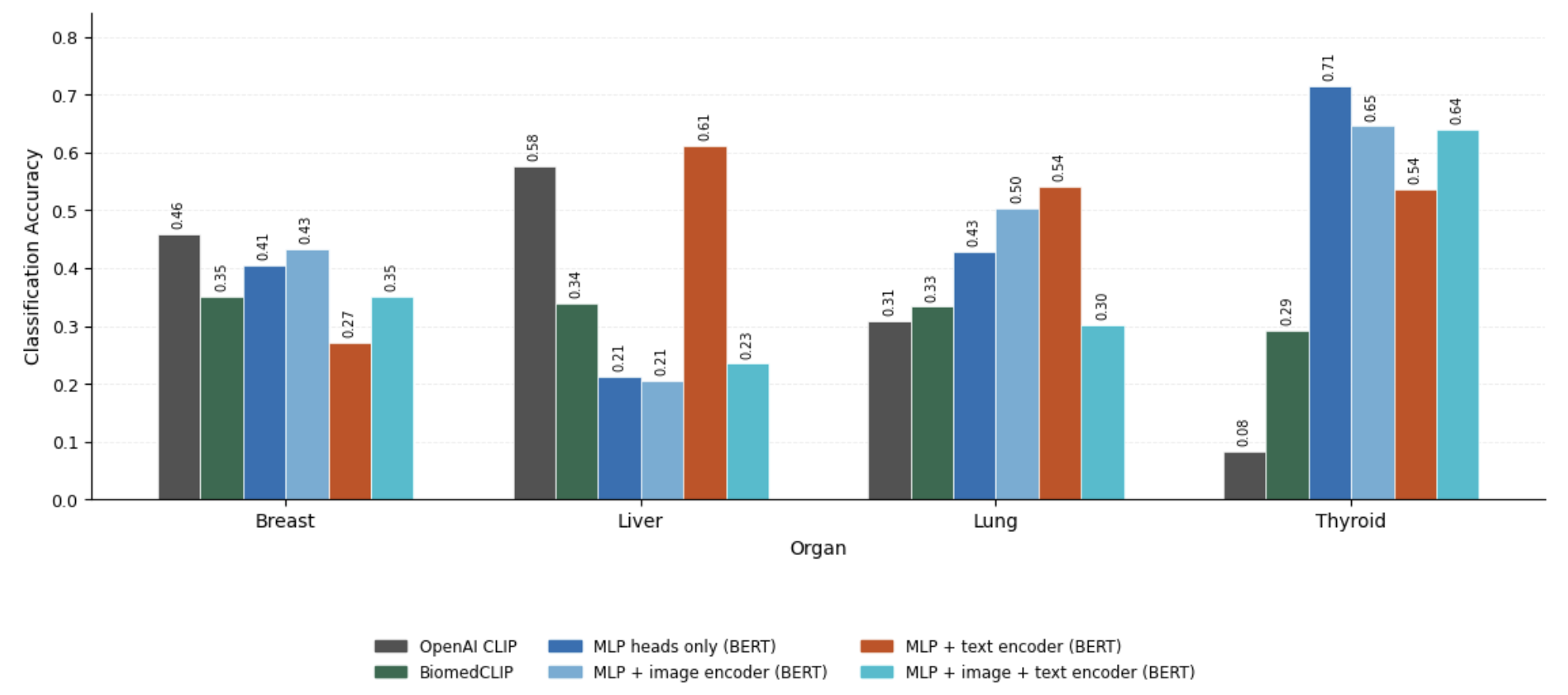}
    \caption{Zero-shot classification accuracy across anatomical domains on the internal held-out test split. Results are shown separately for CLIP-text and BioClinicalBERT-text variants.}
    \label{fig:per-organ-classification}
\end{figure}

To further analyze model behavior, we evaluate classification performance across different anatomical domains (e.g., breast, lung, liver, thyroid). 
This per-organ breakdown reveals substantial heterogeneity in model performance, reflecting both dataset imbalance and varying task difficulty.

In general, models trained with BioClinicalBERT exhibit more stable performance across organs compared to CLIP-text variants. 
This suggests that text encoder trained on clinical language help generalize the models across heterogeneous datasets with different annotation styles and terminology. 
In contrast, CLIP-based models show greater variability: they outperform benchmark models in some domains (e.g., liver) but underperform in others (e.g., breast). 
This pattern indicates that model performance is highly dependent on domain and dataset.

Overall, the per-organ analysis reinforces the main finding from Table~\ref{tab:model_performance}: 
while alignment can be improved through aggressive fine-tuning, 
robust performance and generalization require models to preserve meaningful semantic structure rather than simply maximizing similarity.

\subsection{Qualitative Results}
\begin{figure}[t]
    \centering
    \includegraphics[width=\linewidth]{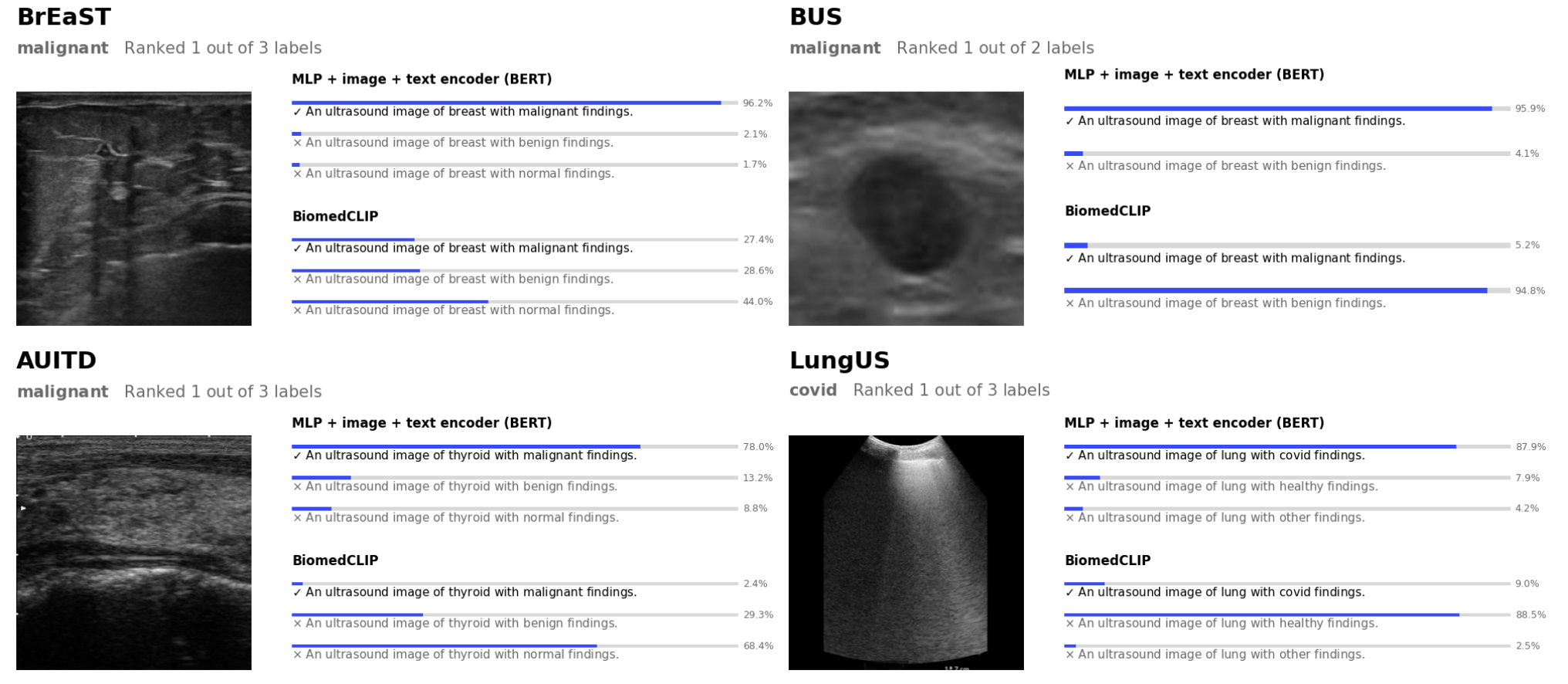}
    \includegraphics[width=\linewidth]{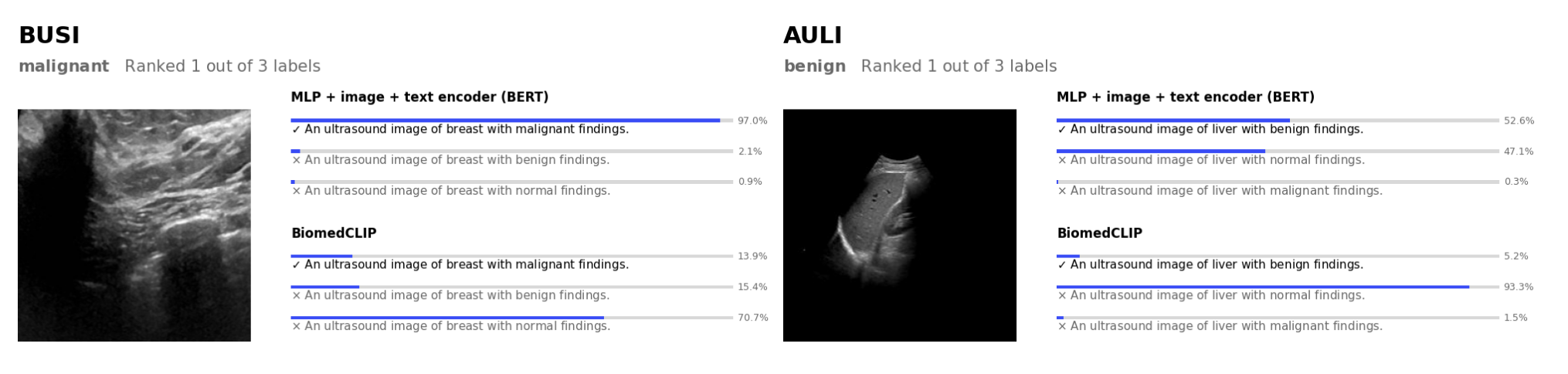}
    \caption{Qualitative comparison of zero-shot predictions from EchoCare-CLIP (BioClinicalBERT text encoder with full fine-tuning) and BiomedCLIP across four internal test datasets and two external held-out datasets.}
    \label{fig:qualitative-result}
\end{figure}

Since the EchoCare-CLIP model using a BioClinicalBERT text encoder with full fine-tuning achieves the best overall performance among all trained variants for the internal test split, we present qualitative examples comparing its predictions with BiomedCLIP. Figure~\ref{fig:qualitative-result} visualizes label ranking outputs for representative cases across four internal test datasets (BrEaST, BUS, AUITD, LungUS) and two external held-out datasets (BUSI, AULI). 

We focus on examples where EchoCare-CLIP outperforms BiomedCLIP, highlighting its ability to assign higher probability to the correct class and produce more discriminative label rankings. These results illustrate how improved alignment can translate into more confident and accurate predictions, particularly in settings where domain-specific training better captures ultrasound-specific visual patterns.

\subsection{Zero-Shot Classification}
\label{subsec:zeroshot}

\begin{figure}[t]
    \centering
    \includegraphics[width=1\linewidth]{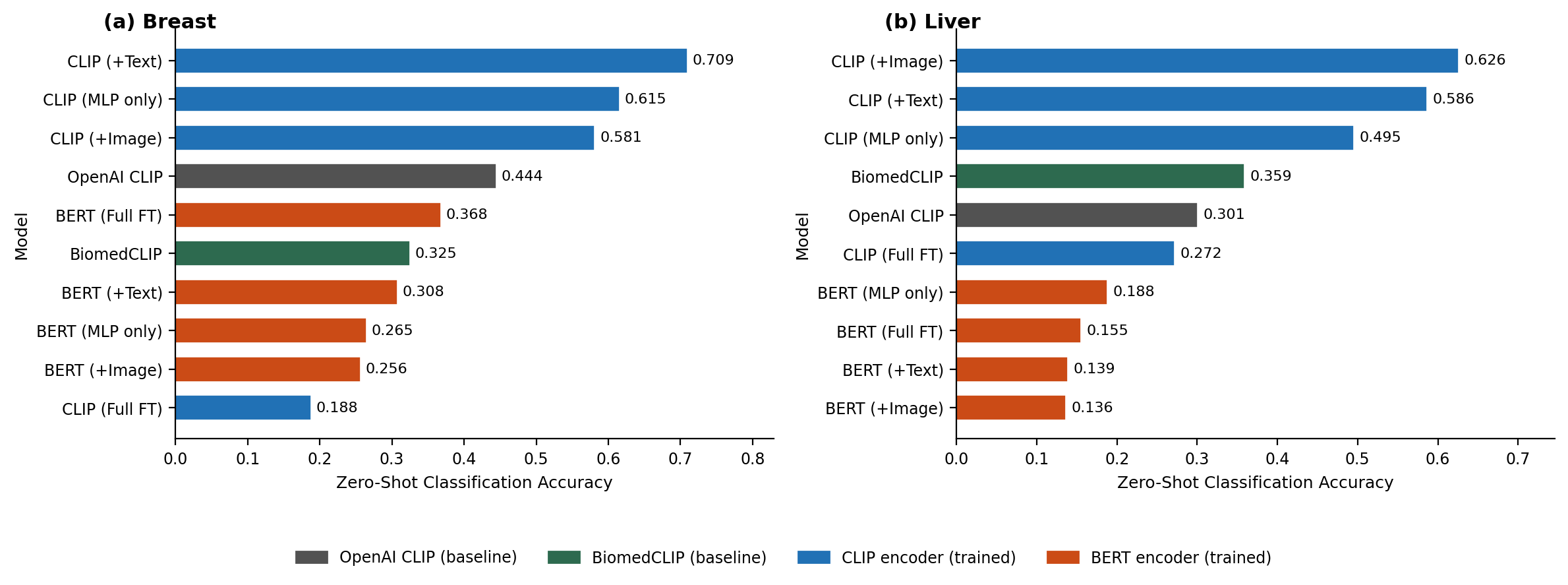}
    \caption{Zero-shot classification accuracy on the BUSI (a) and AULI (b) held-out test sets, with all models ranked from highest to lowest. Colors indicate model family: OpenAI CLIP and BiomedCLIP (baselines), CLIP-encoder variants (blue), and BERT-encoder variants (orange). CLIP-based fine-tuned models dominate the top ranks on BUSI, while performance gaps narrow considerably on AULI.}
    \label{fig:Zero-shot_classification_HO}
\end{figure}

To evaluate zero-shot classification, we compare three prompt template strategies:
(1)~\textbf{Single Template}, in which a single fixed prompt is used for all classes;
(2)~\textbf{Ensemble-Mean}, in which the text embedding for each class is computed as the mean of embeddings produced by ten distinct templates, each incorporating both tissue type and condition labels; 
(3)~\textbf{Ensemble-Max}, in which the alignment score for each class is taken as the maximum across all ten per-class template embeddings.
We found no meaningful difference in zero-shot accuracy across the three strategies (see \S\ref{subsec:caption_templates} for full results), and therefore adopt Single Template for all reported experiments.

The zero-shot evaluations are performed on external held-out datasets (BUSI and AULI), which are not used during training or model selection. As shown in Figure~\ref{fig:Zero-shot_classification_HO}, CLIP-based fine-tuned models occupy the top three positions on both datasets, with CLIP (+Text) achieving the highest accuracy on BUSI (0.709) and CLIP (+Image) leading on AULI (0.626), substantially outperforming the OpenAI CLIP baseline (0.444 and 0.301, respectively). Notably, even the lightest adaptation, training MLP heads only, performs better than both baselines on both datasets (BUSI: 0.615; AULI: 0.495), indicating that domain-specific contrastive training improves text-image alignment even without encoder fine-tuning. BiomedCLIP, despite being pre-trained on biomedical image-text pairs, performs below OpenAI CLIP on BUSI (0.325 vs.\ 0.444) and above it on AULI (0.359
vs.\ 0.301), suggesting its domain advantage is dataset-dependent and does not consistently transfer to ultrasound. The exception among our models is CLIP full fine-tuning, which ranks last on BUSI (0.188) and falls below the OpenAI CLIP baseline on AULI (0.272), suggesting that unconstrained optimization across all parameters degrades the general visual-semantic alignment necessary for zero-shot transfer. BERT-based models consistently occupy the bottom half of the rankings on both datasets,
scoring below all CLIP variants and most baselines, which we attribute to BERT's lack of image-text pre-training: its text representations are not calibrated for cosine similarity-based zero-shot matching.

\subsection{Linear Probe and Few-shot Adaptation}
\label{subsec:linear_fewshot}

\begin{table}[t]
  \caption{Linear probe and few-shot adaptation results on the Breast (BUSI) held-out test set.
           Few-shot AUROC reports mean $\pm$ std across random seeds.}
  \label{tab:results_breast}
  \centering
  \small
  \begin{tabular}{lcccc}
    \toprule
    \textbf{Model}
      & \textbf{Linear Probe}
      & \multicolumn{3}{c}{\textbf{Few-Shot Adaptation (AUROC)}} \\
    \cmidrule(lr){3-5}
      & \textbf{AUROC}
      & \textbf{1\%} & \textbf{5\%} & \textbf{10\%} \\
    \midrule
    \multicolumn{5}{l}{\textit{Baselines}} \\
    OpenAI CLIP  & 0.943          & 0.726{\scriptsize $\pm$0.054} & 0.823{\scriptsize $\pm$0.032} & 0.857{\scriptsize $\pm$0.044} \\
    BiomedCLIP   & \textbf{0.952} & \textbf{0.791}{\scriptsize $\pm$0.049} & \textbf{0.865}{\scriptsize $\pm$0.016} & \textbf{0.892}{\scriptsize $\pm$0.024} \\
    \midrule
    \multicolumn{5}{l}{\textit{CLIP-based}} \\
    \quad MLP heads only    & 0.902 & 0.629{\scriptsize $\pm$0.073} & 0.717{\scriptsize $\pm$0.062} & 0.732{\scriptsize $\pm$0.050} \\
    \quad +Image Enc.       & 0.774 & 0.592{\scriptsize $\pm$0.148} & 0.696{\scriptsize $\pm$0.061} & 0.717{\scriptsize $\pm$0.063} \\
    \quad +Text Enc.        & 0.907 & 0.675{\scriptsize $\pm$0.084} & 0.780{\scriptsize $\pm$0.056} & 0.773{\scriptsize $\pm$0.038} \\
    \quad Full fine-tune    & 0.764 & 0.533{\scriptsize $\pm$0.078} & 0.622{\scriptsize $\pm$0.057} & 0.643{\scriptsize $\pm$0.061} \\
    \midrule
    \multicolumn{5}{l}{\textit{BERT-based}} \\
    \quad MLP heads only    & 0.804 & 0.665{\scriptsize $\pm$0.064} & 0.705{\scriptsize $\pm$0.064} & 0.721{\scriptsize $\pm$0.057} \\
    \quad +Image Enc.       & 0.772 & 0.547{\scriptsize $\pm$0.084} & 0.601{\scriptsize $\pm$0.070} & 0.644{\scriptsize $\pm$0.032} \\
    \quad +Text Enc.        & 0.828 & 0.653{\scriptsize $\pm$0.051} & 0.701{\scriptsize $\pm$0.076} & 0.735{\scriptsize $\pm$0.088} \\
    \quad Full fine-tune    & 0.843 & 0.561{\scriptsize $\pm$0.095} & 0.646{\scriptsize $\pm$0.054} & 0.641{\scriptsize $\pm$0.035} \\
    \bottomrule
  \end{tabular}
\end{table}

\begin{table}[t]
  \caption{Linear probe and few-shot adaptation results on the Liver held-out test set.
           Few-shot AUROC reports mean $\pm$ std across random seeds.}
  \label{tab:results_liver}
  \centering
  \small
  \begin{tabular}{lcccc}
    \toprule
    \textbf{Model}
      & \textbf{Linear Probe}
      & \multicolumn{3}{c}{\textbf{Few-Shot Adaptation (AUROC)}} \\
    \cmidrule(lr){3-5}
      & \textbf{AUROC}
      & \textbf{1\%} & \textbf{5\%} & \textbf{10\%} \\
    \midrule
    \multicolumn{5}{l}{\textit{Baselines}} \\
    OpenAI CLIP  & 0.799 & 0.661{\scriptsize $\pm$0.141} & 0.722{\scriptsize $\pm$0.052} & 0.749{\scriptsize $\pm$0.038} \\
    BiomedCLIP   & 0.729 & 0.657{\scriptsize $\pm$0.064} & 0.724{\scriptsize $\pm$0.027} & 0.760{\scriptsize $\pm$0.028} \\
    \midrule
    \multicolumn{5}{l}{\textit{CLIP-based}} \\
    \quad MLP heads only    & 0.862 & 0.744{\scriptsize $\pm$0.077} & 0.787{\scriptsize $\pm$0.022} & 0.824{\scriptsize $\pm$0.023} \\
    \quad +Image Enc.       & 0.864 & \textbf{0.821}{\scriptsize $\pm$0.024} & \textbf{0.839}{\scriptsize $\pm$0.035} & \textbf{0.864}{\scriptsize $\pm$0.032} \\
    \quad +Text Enc.        & 0.868 & 0.758{\scriptsize $\pm$0.060} & 0.788{\scriptsize $\pm$0.049} & 0.834{\scriptsize $\pm$0.015} \\
    \quad Full fine-tune    & \textbf{0.869} & 0.560{\scriptsize $\pm$0.086} & 0.658{\scriptsize $\pm$0.053} & 0.659{\scriptsize $\pm$0.057} \\
    \midrule
    \multicolumn{5}{l}{\textit{BERT-based}} \\
    \quad MLP heads only    & 0.829 & 0.685{\scriptsize $\pm$0.076} & 0.718{\scriptsize $\pm$0.023} & 0.767{\scriptsize $\pm$0.026} \\
    \quad +Image Enc.       & 0.793 & 0.552{\scriptsize $\pm$0.127} & 0.641{\scriptsize $\pm$0.032} & 0.699{\scriptsize $\pm$0.021} \\
    \quad +Text Enc.        & 0.840 & 0.666{\scriptsize $\pm$0.074} & 0.730{\scriptsize $\pm$0.029} & 0.782{\scriptsize $\pm$0.016} \\
    \quad Full fine-tune    & 0.764 & 0.571{\scriptsize $\pm$0.108} & 0.642{\scriptsize $\pm$0.049} & 0.687{\scriptsize $\pm$0.044} \\
    \bottomrule
  \end{tabular}
\end{table}

Tables~\ref{tab:results_breast} and~\ref{tab:results_liver} report linear
probe AUROC and few-shot adaptation AUROC at 1\%, 5\%, and 10\% label
fractions on BUSI (breast) and AULI (liver), respectively. While the two
datasets differ in absolute performance levels, several patterns hold
consistently across both organs.

The most striking cross-dataset contrast concerns the baselines. On BUSI,
BiomedCLIP achieves the highest linear probe AUROC (0.952) and leads across
all few-shot fractions (0.791 / 0.865 / 0.892), while all of our fine-tuned
models fall below both baselines. On AULI, this ordering reverses sharply:
BiomedCLIP is the weakest model overall (linear probe 0.729), and every
CLIP-based fine-tuned variant surpasses both baselines on linear probe, with
the best reaching 0.869. This reversal suggests that BiomedCLIP's biomedical
pre-training let the image encoder captures features that transfer well to breast tissue but less
so to liver, whereas domain-specific contrastive fine-tuning is more
beneficial where the baseline representations are weaker.

Despite these dataset-level differences, the relative ordering among training
configurations is largely consistent across both organs. The +Text Enc.\
variant is among the strongest fine-tuned models on linear probe for both
BUSI (0.907) and AULI (0.868), and +Image Enc.\ delivers the best few-shot
AUROC on AULI (0.821 / 0.839 / 0.864) while remaining competitive on BUSI.
Full fine-tuning, by contrast, consistently underperforms on few-shot
adaptation on both datasets, achieving only 0.533 on BUSI and 0.560 on
AULI at 1\%, despite attaining a competitive or best linear probe score
(0.764 on BUSI; 0.869 on AULI). This trade-off, consistent across organs,
reflects the overfitting risk of unconstrained end-to-end training: the model
fits the contrastive objective closely but loses the generalizable
unimodal image representations needed for low-data transfer. CLIP +Image Enc.\ also exhibits
notably high variance on BUSI at 1\% ({\small$\pm$}0.148), suggesting
instability under extreme data scarcity when the visual encoder is unfrozen.

BERT-based models show consistently weaker performance than their CLIP-based
counterparts across both datasets. On BUSI, the best BERT model (full
fine-tune, AUROC 0.843) sits 11 points below BiomedCLIP on linear probe and
trails all CLIP-based variants on few-shot. The same ordering holds on AULI,
where BERT variants generally fail to exceed the OpenAI CLIP baseline on
few-shot despite modest linear probe gains. This consistent gap across two
anatomically distinct organs points to a structural limitation: without
image-text pre-training, Bio-ClinicalBERT text representations do not provide
the aligned embedding geometry that benefits the image encoder during contrastive training.

\section{Ablation Study}
\subsection{Template vs. LLM Caption Expansion}
\begin{figure}[t]
    \centering
    \includegraphics[width=1\linewidth]{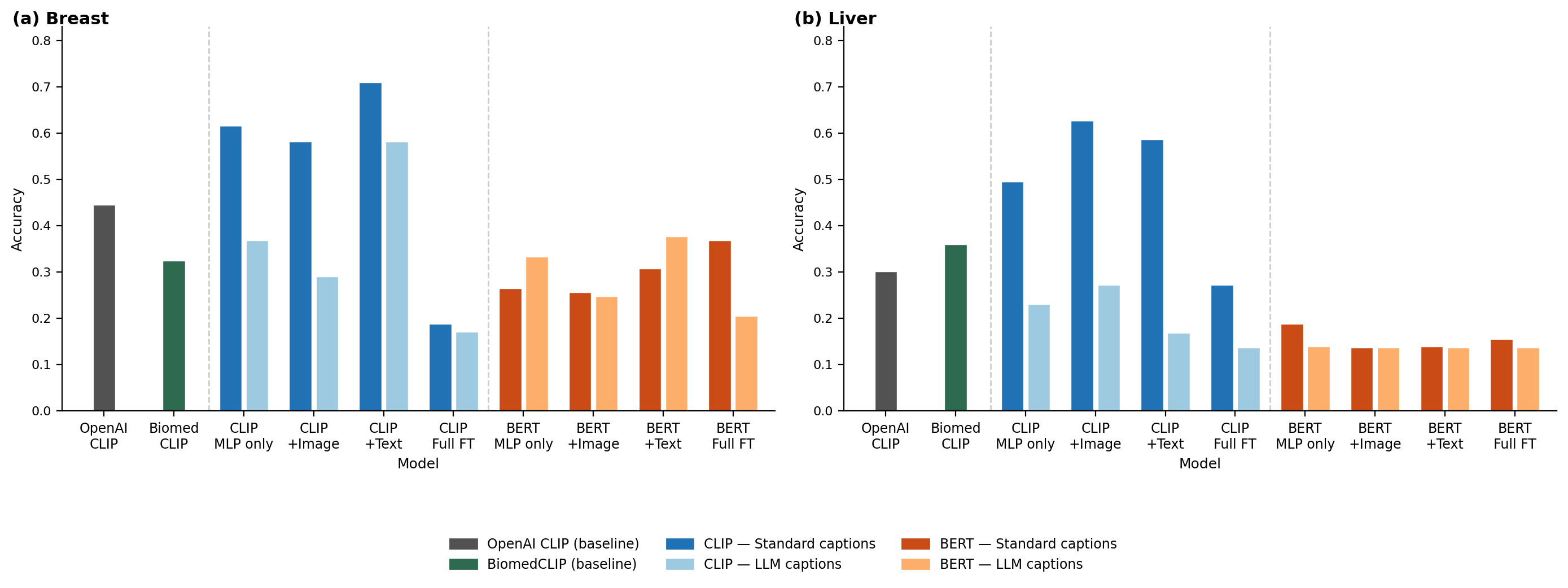}
    \caption{Zero-shot classification accuracy on the BUSI (a) and AULI (b) held-out test sets, grouped by model architecture. Within each group, dark bars show results with standard captions and light bars with LLM-generated captions. CLIP-based models generally benefit more from fine-tuning than BERT-based models, and standard captions tend to match or outperform LLM captions across most configurations.}
    \label{fig:Zero-Shot_Classification_Template_LLM_Comparision}
\end{figure}

LLM-generated captions do not yield a consistent benefit over template-based captions across our experiments. As shown in Figure~\ref{fig:Zero-Shot_Classification_Template_LLM_Comparision}, for the stronger CLIP-based configurations, LLM captions actually degraded zero-shot classification performance relative to their template-based counterparts. Although LLM generation substantially increased lexical complexity — expanding caption length, unigram and bigram diversity, and Distinct-$n$ scores (see Extended Materials Section~\ref{subsec:llm_based_caption_generation}) - the underlying information content did not change, as both caption types are derived from the same structured metadata. The clinically meaningful signal available to the model therefore remains the same regardless of surface-form variation.

We hypothesize two complementary explanations for this result. First, without access to the image itself, the language model cannot introduce genuinely new clinical observations; paraphrasing existing metadata fields produces linguistic variation without semantic enrichment. Second, template-based captions share a more consistent vocabulary with the short label prompts used at evaluation time, which may reduce the distributional gap between training captions and inference-time text queries. Taken together, these findings suggest that increased lexical complexity is not a proxy for caption quality in contrastive vision-language training.

\section{Discussion}
\label{sec:discussion}

In this work, we developed an ultrasound vision-language alignment framework by combining an ultrasound-specific EchoCare image encoder with either a CLIP text encoder or BioClinicalBERT text encoder. Across experiments, contrastive training improved image-text alignment relative to off-the-shelf baselines, but the relationship between alignment and downstream performance was not straightforward. Models with more trainable components generally achieved higher paired alignment scores, however, stronger alignment did not always translate into better zero-shot classification, linear probing, or few-shot adaptation. This suggests that optimizing for cross-modal similarity can overfit the caption distribution without necessarily preserving important class-discriminative visual representations.

One key finding is that encoder fine-tuning reduces the ultrasound domain gap but introduces overfitting risk. Fine-tuning the image encoder helped adapt the representation space to ultrasound-specific visual patterns, particularly in organs where benchmark models performed poorly. At the same time, full end-to-end fine-tuning often degraded transfer performance, especially in few-shot settings. This pattern suggests that aggressive fine-tuning may distort pretrained representations and reduce their generalizability. In practice, partial adaptation strategies may offer a better trade-off between domain adaptation and robustness.

We also found that the choice of text encoder affects generalization in different ways. 
BioClinicalBERT-based models achieved stronger performance on the internal held-out test split, suggesting that clinical language priors can help capture medically relevant semantics within the training distribution. 
However, CLIP-text variants performed better on completely held-out datasets for zero-shot classification, linear probing, and few-shot adaptation. 
This suggests that the BioClinicalBERT models may be more prone to fitting dataset-specific caption patterns, probably because the BERT-based configurations introduce more trainable parameters. 
In contrast, the CLIP text encoder already has a representation space optimized for cosine-similarity-based image-text matching, which may make it more robust under distribution shift. 
Overall, these results suggest a trade-off: clinical text encoders may improve in-domain semantic fit, while CLIP-style text encoders may preserve stronger generalizable alignment for transfer tasks.

Another important observation is that caption consistency may matter more than linguistic diversity for zero-shot transfer. Although LLM-generated captions increased text diversity and produced more natural clinical language, this did not consistently improve zero-shot performance. One explanation is that contrastive models benefit from stable and repeated semantic anchors, especially when the training corpus is modest. Highly diverse captions may introduce only surface-level linguistic variation without adding new visual information, making the alignment task noisier. This suggests that future caption generation should prioritize clinically consistent, structured descriptions rather than more diversity.

Our study has several limitations. First, the training corpus remains modest compared with large-scale vision-language pretraining datasets. Although our dataset spans multiple organs and includes expert-derived reports, its scale is still far smaller than general CLIP or biomedical CLIP corpora. This limits the ability of the model to learn robust cross-organ and cross-dataset representations, especially for rare labels or anatomical domains.

Second, organ-level class imbalance likely affects both training and evaluation. Some organs and diagnostic categories have substantially more samples than others, which can bias the contrastive objective toward dominant domains. This imbalance also complicates interpretation of aggregate metrics, since strong performance on large or easier datasets may hide weaker performance on smaller or more clinically ambiguous subsets. Our per-organ results show that model behavior varies substantially across anatomical domains, reinforcing the need for more balanced evaluation.

Third, we did not perform exhaustive hyperparameter optimization for every configuration. Because each model variant differs in the number of trainable parameters and in the encoder architecture, the same learning rate, batch size, or temperature setting may not be optimal across all settings. Future work should tune hyperparameters separately for each fine-tuning regime.

Finally, our retrieval evaluation is limited by imperfect image--report pairing in the full held-out test split (see ~\ref{subsec:retrieval}). Although most samples are expected to have meaningful image-text correspondence, strict one-to-one pairing is not guaranteed for all cases. This can introduce noise into Recall@K metrics and underestimate retrieval performance. A curated retrieval benchmark with a subset of verified one-to-one image--report pairs would provide a better assessment of cross-modal retrieval.

Overall, our results suggest that ultrasound vision-language modeling requires more than simply applying CLIP-style training to ultrasound images. Domain adaptation improves alignment, but robust clinical transfer depends on preserving discriminative visual structure, using clinically grounded text representations, and ensuring high-quality image-text supervision. These findings motivate future work on larger curated datasets, better-balanced organ sampling, more controlled caption generation, and training objectives that jointly optimize alignment, retrieval, and diagnostic discrimination.

\section{Conclusion}
\label{sec:conclusion}

Ultrasound foundation models have achieved strong performance on structured
prediction tasks but remain vision-only, precluding zero-shot and few-shot
transfer without task-specific annotation. We addressed this gap with
EchoCare-CLIP, a CLIP-style dual-encoder framework that aligns ultrasound
images with clinical text through symmetric InfoNCE contrastive training.
Across eight fine-tuning configurations spanning two text encoder families and
two caption strategies, our models consistently improve cross-modal alignment
over off-the-shelf baselines. However, improved alignment does not necessarily yield stronger uni-modal
representations. While contrastive training increases image-text similarity,
it can degrade classification performance due to overfitting to caption
structure or image feature distribution. Similarly, encoder fine-tuning reduces the domain gap but introduces
overfitting risk and degrades quality of output representation, with partial fine-tuning offering a better balance between
alignment and generalization. On external held-out datasets, CLIP-based
variants achieve strong zero-shot classification accuracy (up to 0.709 on BUSI
and 0.626 on AULI) and competitive few-shot AUROC, with partial fine-tuning
strategies offering the best balance between linear separability and low-data
generalization. We further show that template-based captions match or
outperform LLM-generated captions, suggesting that caption consistency is more
important than lexical diversity for contrastive ultrasound pretraining.

Several directions remain open. Scaling the training corpus, particularly
with more balanced organ representation, would likely improve cross-organ
robustness. Per-configuration hyperparameter tuning and evaluation against
finer-grained clinical taxonomies such as BI-RADS and LI-RADS would better
characterize clinical utility. Benchmarking directly against
EchoCare~\cite{zhang2025echocare} and USFM~\cite{jiao2024usfm} on shared
tasks would further contextualize where vision-language alignment adds value
over vision-only pretraining.
\section*{Acknowledgments}

This work was completed as part of BMI 702 at Harvard. We thank the BMI 702 course staff for guidance throughout the project, and extend special thanks to Teaching Fellow Mohammed Baharoon for his mentorship and invaluable feedback. We gratefully acknowledge the authors of EchoCare~\cite{zhang2025echocare} for publicly releasing their pretrained SwinTransformerV2~\cite{liu2022swinv2} weights, which served as our image encoder backbone. Pretrained weights for OpenAI CLIP~\cite{radford2021clip,clipvitbasepatch32_hf} and BiomedCLIP~\cite{zhang2025biomedclipmultimodalbiomedicalfoundation,biomedclip_hf} were accessed via the Hugging Face model hub. LLM-based caption generation used Qwen3-4B-Instruct-2507~\cite{Qwen3-4B-Instruct-2507}, also accessed via Hugging Face. We made use of several publicly available datasets: BUSI~\cite{al-dhabyani2020busi}, AULI~\cite{xu2022auli}, AUITD~\cite{azouz2023auitd}, BrEaST~\cite{pawlowska2024breast}, the Lung Ultrasound
dataset~\cite{katumba2025lung}, BUS~\cite{yu2026buscot}, and the Report-Gen Thyroid and Liver datasets~\cite{li2024usreport}.

\bibliographystyle{unsrt}
\bibliography{references}

@InProceedings{radford2021clip,
  title = 	 {Learning Transferable Visual Models From Natural Language Supervision},
  author =       {Radford, Alec and Kim, Jong Wook and Hallacy, Chris and Ramesh, Aditya and Goh, Gabriel and Agarwal, Sandhini and Sastry, Girish and Askell, Amanda and Mishkin, Pamela and Clark, Jack and Krueger, Gretchen and Sutskever, Ilya},
  booktitle = 	 {Proceedings of the 38th International Conference on Machine Learning},
  pages = 	 {8748--8763},
  year = 	 {2021},
  editor = 	 {Meila, Marina and Zhang, Tong},
  volume = 	 {139},
  series = 	 {Proceedings of Machine Learning Research},
  month = 	 {18--24 Jul},
  publisher =    {PMLR},
  url = 	 {https://proceedings.mlr.press/v139/radford21a.html}
}

@article{zhang2025echocare,
  title={A Fully Open and Generalizable Foundation Model for Ultrasound Clinical Applications},
  author={Zhang, Hongyuan and Wu, Yuheng and Zhao, Mingyang and Chen, Zhiwei and Li, Rebecca and Zhu, Fei and Zhao, Haohan and Yuan, Xiaohua and Yang, Meng and Qiu, Chunli and others},
  journal={arXiv preprint arXiv:2509.11752},
  year={2025}
}

@article{jiao2024usfm,
title = {USFM: A universal ultrasound foundation model generalized to tasks and organs towards label efficient image analysis},
journal = {Medical Image Analysis},
volume = {96},
pages = {103202},
year = {2024},
issn = {1361-8415},
doi = {https://doi.org/10.1016/j.media.2024.103202},
url = {https://www.sciencedirect.com/science/article/pii/S1361841524001270},
author = {Jing Jiao and Jin Zhou and Xiaokang Li and Menghua Xia and Yi Huang and Lihong Huang and Na Wang and Xiaofan Zhang and Shichong Zhou and Yuanyuan Wang and Yi Guo}
}

@article{wells2006ultrasound,
  title   = {Ultrasound Imaging},
  author  = {Wells, Peter N. T.},
  journal = {Physics in Medicine \& Biology},
  year    = {2006},
  volume  = {51},
  number  = {13},
  pages   = {R83--R98},
  doi     = {10.1088/0031-9155/51/13/R01}
}

@article{Strauss2007inter_intra_variability_sonograph,
  author       = {Strauss, S. and Gavish, E. and Gottlieb, P. and Katsnelson, L.},
  title        = {Interobserver and Intraobserver Variability in the Sonographic Assessment of Fatty Liver},
  journal      = {AJR American Journal of Roentgenology},
  year         = {2007},
  volume       = {189},
  number       = {6},
  pages        = {W320--W323},
  doi          = {10.2214/AJR.07.2123},
}

@article{Itani2019inter_variability_sonograph,
  author       = {Itani, Malak and Assaker, Richard and Moshiri, Mariam and Dubinsky, Theodore J. and Dighe, Manjiri K.},
  title        = {Inter-observer Variability in the American College of Radiology Thyroid Imaging Reporting and Data System: In-Depth Analysis and Areas for Improvement},
  journal      = {Ultrasound in Medicine and Biology},
  year         = {2019},
  volume       = {45},
  number       = {2},
  pages        = {461--470},
  doi          = {10.1016/j.ultrasmedbio.2018.09.026},
}

@article{Zhang2020convirt,
  author       = {Zhang, Yuhao and Jiang, Hang and Miura, Yasuhide and Manning, Christopher D. and Langlotz, Curtis P.},
  title        = {Contrastive Learning of Medical Visual Representations from Paired Images and Text},
  journal      = {arXiv preprint},
  year         = {2020},
  note         = {arXiv:2010.00747},
  doi          = {10.48550/arXiv.2010.00747},
}

@article{Lu2024conch,
  author       = {Lu, Ming Y. and Chen, Bowen and Williamson, Drew F. K. and Chen, Richard J. and Liang, Ivy and Ding, Tong and Jaume, Guillaume and Odintsov, Igor and Le, Long Phi and Gerber, Georg and Parwani, Anil V. and Zhang, Andrew and Mahmood, Faisal},
  title        = {A visual-language foundation model for computational pathology},
  journal      = {Nature Medicine},
  year         = {2024},
  volume       = {30},
  number       = {3},
  pages        = {863--874},
  doi          = {10.1038/s41591-024-02856-4},
}

@article{li2024usreport,
  title   = {Ultrasound Report Generation with Cross-Modality Feature Alignment via Unsupervised Guidance},
  author  = {Li, Jun and Su, Tongkun and Zhao, Baoliang and Lv, Faqin and Wang, Qiong and Navab, Nassir and Hu, Ying and Jiang, Zhongliang},
  journal = {IEEE Transactions on Medical Imaging},
  year    = {2024},
  doi     = {10.1109/TMI.2024.3424978}
}

@article{al-dhabyani2020busi,
  author  = {Al-Dhabyani, Wafaa and Gomaa, Mohammed and Khaled, Hussien and Fahmy, Aly},
  title   = {Dataset of breast ultrasound images},
  journal = {Data in Brief},
  volume  = {28},
  pages   = {104863},
  year    = {2020},
  doi     = {10.1016/j.dib.2019.104863}
}

@article{pawlowska2024breast,
  author  = {Paw{\l}owska, Anna and {\'C}wierz-Pie{\'n}kowska, Anna and Domalik, Aleksandra and Jagu{\'s}, Dominika and Kasprzak, Piotr and Matkowski, Rafa{\l} and Fura, {\L}ukasz and Nowicki, Andrzej and {\.Z}o{\l}ek, Natalia},
  title   = {Curated benchmark dataset for ultrasound based breast lesion analysis},
  journal = {Scientific Data},
  volume  = {11},
  number  = {1},
  pages   = {148},
  year    = {2024},
  doi     = {10.1038/s41597-024-02984-z}
}

@article{katumba2025lung,
  author  = {Katumba, Allan and Murindanyi, Samuel and Okila, Nicolas and Nakatumba-Nabende, Joyce and Mwikirize, Christopher and Serugunda, Joseph and Bugeza, Stephen and Oriekot, Amon and Bossa, Joseph and Nabawanuka, Esther},
  title   = {A dataset of lung ultrasound images for automated AI-based lung disease classification},
  journal = {Data in Brief},
  volume  = {62},
  pages   = {112034},
  year    = {2025},
  doi     = {10.1016/j.dib.2025.112034}
}

@misc{xu2022auli,
  author       = {Xu, Yonghao and Zheng, Bing and Liu, Xiaoyu and Wu, Tianfu and Ju, Jia and Wang, Shuo and Lian, Yijun and Zhang, Hong and Liang, Tian and Sang, Yang and Jiang, Rui and Wang, Guang and Ren, Jianfeng and Chen, Tianfu},
  title        = {Annotated Ultrasound Liver images},
  year         = {2022},
  publisher    = {Zenodo},
  doi          = {10.5281/zenodo.7272660},
  url          = {https://zenodo.org/records/7272660}
}

@misc{azouz2023auitd,
  author       = {Azouz, Maroua},
  title        = {Algerian Ultrasound Images Thyroid Dataset: AUITD},
  year         = {2023},
  publisher    = {Kaggle},
  url          = {https://www.kaggle.com/datasets/azouzmaroua/algeria-ultrasound-images-thyroid-dataset-auitd},
  note         = {Accessed: 2026-03-17}
}

@misc{zhang2025biomedclipmultimodalbiomedicalfoundation,
      title={BiomedCLIP: a multimodal biomedical foundation model pretrained from fifteen million scientific image-text pairs}, 
      author={Sheng Zhang and Yanbo Xu and Naoto Usuyama and Hanwen Xu and Jaspreet Bagga and Robert Tinn and Sam Preston and Rajesh Rao and Mu Wei and Naveen Valluri and Cliff Wong and Andrea Tupini and Yu Wang and Matt Mazzola and Swadheen Shukla and Lars Liden and Jianfeng Gao and Angela Crabtree and Brian Piening and Carlo Bifulco and Matthew P. Lungren and Tristan Naumann and Sheng Wang and Hoifung Poon},
      year={2025},
      eprint={2303.00915},
      archivePrefix={arXiv},
      primaryClass={cs.CV},
      url={https://arxiv.org/abs/2303.00915}, 
}

@inproceedings{liu2022swinv2,
  title     = {Swin Transformer {V2}: Scaling Up Capacity and Resolution},
  author    = {Liu, Ze and Hu, Han and Lin, Yutong and Yao, Zhuliang and Xie, Zhenda and Wei, Yixuan and Ning, Jia and Cao, Yue and Zhang, Zheng and Dong, Li and Wei, Furu and Guo, Baining},
  booktitle = {Proceedings of the IEEE/CVF Conference on Computer Vision and Pattern Recognition (CVPR)},
  year      = {2022},
  doi       = {10.48550/arXiv.2111.09883}
}

@inproceedings{loshchilov2019adamw,
  title     = {Decoupled Weight Decay Regularization},
  author    = {Loshchilov, Ilya and Hutter, Frank},
  booktitle = {International Conference on Learning Representations (ICLR)},
  year      = {2019},
  url       = {https://openreview.net/forum?id=Bkg6RiCqY7}
}

@article{oord2018infonce,
  title   = {Representation Learning with Contrastive Predictive Coding},
  author  = {Oord, Aaron van den and Li, Yazhe and Vinyals, Oriol},
  journal = {arXiv preprint arXiv:1807.03748},
  year    = {2018},
  url     = {https://arxiv.org/abs/1807.03748}
}

@inproceedings{loshchilov2017sgdr,
  title     = {SGDR: Stochastic Gradient Descent with Warm Restarts},
  author    = {Loshchilov, Ilya and Hutter, Frank},
  booktitle = {International Conference on Learning Representations (ICLR)},
  year      = {2017},
  url       = {https://openreview.net/forum?id=Skq89Scxx}
}

@article{johnson2019mimiccxr,
  title   = {MIMIC-CXR, a de-identified publicly available database of chest radiographs with free-text reports},
  author  = {Johnson, Alistair E. W. and Pollard, Tom J. and Berkowitz, Seth J. and Greenbaum, Nathaniel R. and Lungren, Matthew P. and Deng, Chih-ying and Mark, Roger G. and Horng, Steven},
  journal = {Scientific Data},
  volume  = {6},
  pages   = {317},
  year    = {2019},
  publisher = {Nature Publishing Group},
  doi     = {10.1038/s41597-019-0322-0}
}

@misc{clipvitbasepatch32_hf,
  title  = {CLIP ViT-B/32},
  author = {OpenAI},
  year   = {2021},
  howpublished = {\url{https://huggingface.co/openai/clip-vit-base-patch32}},
  note   = {Accessed: 2026-04-17}
}

@misc{biomedclip_hf,
  author       = {Sheng Zhang and Yanbo Xu and Naoto Usuyama and others},
  title        = {{BiomedCLIP-PubMedBERT\_256-vit\_base\_patch16\_224}},
  year         = {2024},
  howpublished = {\url{https://huggingface.co/microsoft/BiomedCLIP-PubMedBERT_256-vit_base_patch16_224}},
  note         = {Hugging Face model repository}
}

@article{jiang2023ultrasound_med_background,
  title     = {Robotic Ultrasound Imaging: State-of-the-Art and Future Perspectives},
  author    = {Jiang, Zhongliang and Salcudean, Septimiu E. and Navab, Nassir},
  journal   = {Medical Image Analysis},
  volume    = {89},
  pages     = {102878},
  year      = {2023},
  month     = {October},
  publisher = {Elsevier},
  doi       = {10.1016/j.media.2023.102878},
  pmid      = {37541100}
}

@article{litjens2017survey_med_image,
  title     = {A Survey on Deep Learning in Medical Image Analysis},
  author    = {Litjens, Geert and Kooi, Thijs and Bejnordi, Babak Ehteshami and
               Setio, Arnaud Arindra Adiyoso and Ciompi, Francesco and
               Ghafoorian, Mohsen and van der Laak, Jeroen A. W. M. and
               van Ginneken, Bram and S{\'a}nchez, Clara I.},
  journal   = {Medical Image Analysis},
  volume    = {42},
  pages     = {60--88},
  year      = {2017},
  doi       = {10.1016/j.media.2017.07.005},
  pmid      = {28778026}
}

@misc{Qwen3-4B-Instruct-2507,
  author = {Qwen Team},
  title  = {Qwen3-4B-Instruct-2507},
  year   = {2025},
  publisher = {Hugging Face},
  journal = {Hugging Face Repository},
  howpublished = {\url{https://huggingface.co/Qwen/Qwen3-4B-Instruct-2507}}
}

@inproceedings{alsentzer2019bioclinicalbert,
  title     = {Publicly Available Clinical {BERT} Embeddings},
  author    = {Alsentzer, Emily and Murphy, John R. and Boag, Willie and
               Weng, Wei-Hung and Jin, Di and Naumann, Tristan and
               McDermott, Matthew B. A.},
  booktitle = {Proceedings of the 2nd Clinical Natural Language Processing
               Workshop (Clinical NLP) at NAACL},
  year      = {2019},
  url       = {https://arxiv.org/abs/1904.03323}
}

@misc{alsentzer2019bioclinicalbert_hf,
  author    = {Alsentzer, Emily},
  title     = {{Bio\_ClinicalBERT}},
  year      = {2019},
  publisher = {Hugging Face},
  howpublished = {\url{https://huggingface.co/emilyalsentzer/Bio_ClinicalBERT}},
  note      = {Accessed: 2026-04-29}
}

@article{yu2026buscot,
  title   = {A Chain-of-thought Reasoning Breast Ultrasound Dataset Covering All Histopathology Categories},
  author  = {Yu, Haojun and Li, Youcheng and Niu, Zihan and Zhang, Nan and Gong, Xuantong and Li, Huan and Zou, Zhiying and Qi, Haifeng and Cao, Zhenxiao and Lan, Zijie and Yuan, Xingjian and He, Jiating and Zhang, Haokai and Zhang, Shengtao and Wang, Zicheng and Wang, Dong and Zhao, Ziwei and Chen, Congying and Wang, Yong and Qin, Wangyan and Zhu, Qingli and Wang, Liwei},
  journal = {Scientific Data},
  volume  = {13},
  pages   = {370},
  year    = {2026},
  doi     = {10.1038/s41597-026-06702-9}
}

\section*{Extended Materials}
\captionsetup[table]{name=Extended Table}

\captionsetup[figure]{name=Extended Figure}

\setcounter{table}{0}

\setcounter{figure}{0}

\subsection{Caption Templates}
\label{subsec:caption_templates}

Extended Table~\ref{tab:caption_templates} lists all 30 caption templates used in the three-tier caption generation pipeline. Placeholders are typeset in \texttt{monospace} and are resolved at runtime from dataset metadata; default fallback strings are shown in Extended Table~\ref{tab:placeholders}.

\begin{longtable}{cp{12cm}}
    \caption{Caption template library.}
    \label{tab:caption_templates} \\

    \toprule
    \textbf{\#} & \textbf{Template} \\
    \midrule
    \endfirsthead

    \multicolumn{2}{l}{\footnotesize\textit{(continued from previous page)}} \\
    \toprule
    \textbf{\#} & \textbf{Template} \\
    \midrule
    \endhead

    \midrule
    \multicolumn{2}{r}{\footnotesize\textit{(continued on next page)}} \\
    \endfoot

    \bottomrule
    \endlastfoot

    \multicolumn{2}{l}{\textit{Tier 1 — Tissue label only
                               (placeholder: \texttt{\{Tissue\}})}} \\
    \midrule
    1  & An ultrasound image of \texttt{\{Tissue\}}. \\
    2  & A B-mode ultrasound showing \texttt{\{Tissue\}}. \\
    3  & Sonographic appearance of \texttt{\{Tissue\}}. \\
    4  & This is an ultrasound image of \texttt{\{Tissue\}}. \\
    5  & A grayscale ultrasound image demonstrating \texttt{\{Tissue\}}. \\
    6  & Ultrasound findings consistent with \texttt{\{Tissue\}}. \\
    7  & A clinical ultrasound scan of \texttt{\{Tissue\}}. \\
    8  & An echographic image of \texttt{\{Tissue\}}. \\
    9  & This sonogram shows \texttt{\{Tissue\}}. \\
    10 & A diagnostic ultrasound image of \texttt{\{Tissue\}},
         obtained for clinical evaluation. \\
    \midrule

    \multicolumn{2}{l}{\textit{Tier 2 — Tissue and condition labels
                               (placeholders: \texttt{\{Tissue\}},
                               \texttt{\{Condition\}})}} \\
    \midrule
    1  & An ultrasound image of \texttt{\{Tissue\}} with
         \texttt{\{Condition\}} findings. \\
    2  & A B-mode ultrasound of \texttt{\{Tissue\}}, consistent with
         \texttt{\{Condition\}}. \\
    3  & Sonographic appearance of \texttt{\{Condition\}}
         \texttt{\{Tissue\}}. \\
    4  & This ultrasound demonstrates \texttt{\{Tissue\}} exhibiting
         features of \texttt{\{Condition\}}. \\
    5  & A clinical ultrasound scan of \texttt{\{Tissue\}}, indicative
         of \texttt{\{Condition\}} pathology. \\
    6  & Echographic findings of \texttt{\{Tissue\}} showing
         \texttt{\{Condition\}} characteristics. \\
    7  & This sonogram of \texttt{\{Tissue\}} is consistent with a
         \texttt{\{Condition\}} diagnosis. \\
    8  & A diagnostic ultrasound image of \texttt{\{Tissue\}}, with
         imaging features suggestive of \texttt{\{Condition\}}. \\
    9  & Ultrasound of \texttt{\{Tissue\}} presenting sonographic signs
         of \texttt{\{Condition\}}. \\
    10 & A grayscale ultrasound demonstrating \texttt{\{Condition\}}
         changes in \texttt{\{Tissue\}}. \\
    \midrule

    \multicolumn{2}{l}{\textit{Tier 3 — Metadata (placeholders
                               see Extended Table~\ref{tab:placeholders})}} \\
    \midrule
    1  & Ultrasound of \texttt{\{Region\}} in \texttt{\{PatientInfo\}}.
         \texttt{\{Findings\}}. Assessment: \texttt{\{Condition\}}. \\
    2  & Sonographic findings in \texttt{\{Region\}}:
         \texttt{\{Findings\}}. \texttt{\{PatientInfo\}}.
         Diagnosis: \texttt{\{Condition\}}. \\
    3  & An ultrasound image of \texttt{\{Region\}} consistent with
         \texttt{\{Condition\}}. \texttt{\{Findings\}}. \\
    4  & \texttt{\{Region\}} evaluated by sonography.
         \texttt{\{Findings\}}. Conclusion: \texttt{\{Condition\}}. \\
    5  & \texttt{\{PatientInfo\}} underwent ultrasound imaging.
         \texttt{\{Findings\}} identified in \texttt{\{Region\}},
         suggestive of \texttt{\{Condition\}}. \\
    6  & Sonography: \texttt{\{Region\}}, \texttt{\{PatientInfo\}}.
         \texttt{\{Findings\}}. \texttt{\{Condition\}}. \\
    7  & \texttt{\{PatientInfo\}}. Ultrasound of \texttt{\{Region\}}
         reveals \texttt{\{Findings\}}, consistent with
         \texttt{\{Condition\}}. \\
    8  & \texttt{\{Condition\}} pattern on ultrasound.
         \texttt{\{Region\}} demonstrates \texttt{\{Findings\}}.
         \texttt{\{PatientInfo\}}. \\
    9  & Sonographic examination of \texttt{\{Region\}}.
         \texttt{\{Findings\}}. \texttt{\{Condition\}}. \\
    10 & Ultrasound performed on \texttt{\{PatientInfo\}}.
         Examination of \texttt{\{Region\}} revealed
         \texttt{\{Findings\}}. Impression: \texttt{\{Condition\}}. \\
\end{longtable}

\begin{table}[h]
    \centering
    \caption{Tier 3 placeholder definitions: source metadata fields and
             fallback strings used when a field is absent.}
    \label{tab:placeholders}
    \begin{tabular}{lp{5.5cm}l}
        \toprule
        \textbf{Placeholder} & \textbf{Example source columns} &
        \textbf{Default (missing)} \\
        \midrule
        \texttt{\{PatientInfo\}} & Age, Gender, Pregnancy status &
            ``a patient'' \\
        \texttt{\{Region\}} & Tissue, Tissue composition, Zone &
            ``unknown region'' \\
        \texttt{\{Findings\}} & Shape, Margin, Echogenicity,
            Posterior features, Consolidation, Effusion &
            ``unremarkable findings'' \\
        \texttt{\{Condition\}} & Diagnosis, Classification, BI-RADS,
            Interpretation, Zone-derived severity &
            ``unspecified condition'' \\
        \bottomrule
    \end{tabular}
\end{table}

\subsection{LLM-Based Caption Generation Detail}
\label{subsec:llm_based_caption_generation}

We utilized the \textsc{Qwen3-4B-Instruct-2507}\cite{Qwen3-4B-Instruct-2507} model with a temperature of 1.0, a top-p value of 0.95, and a repetition penalty of 1.05.

The system prompt used for generation is provided below in ~\ref{quote:llm_prompt}.

\begin{quote}
\label{quote:llm_prompt}
\ttfamily\small\setlength{\parskip}{0.5em}
You are a radiology assistant writing English captions for ultrasound images.
You are given structured metadata fields (key: value pairs) from a clinical dataset.
Produce THREE DIFFERENT captions for the SAME image.

Content rules (apply to every caption):
- Ground the caption ONLY in the provided fields. Never invent values, measurements, or findings.
- Incorporate EVERY provided field (diagnosis, BI-RADS, pathology, anatomical region, lesion descriptors, patient demographics, findings, etc.). Do not omit any.
- Read like natural clinical language written by a radiologist, not a key-value dump.
- Length: 15 to 60 words, one to three sentences. Length should scale with the number of provided fields: shorter captions (~15-25 words) when few fields are given, and longer captions (up to ~40-60 words, two or more sentences) when many fields are given, so that every field is naturally incorporated.

Diversity rules (across the three captions):
- Captions must differ meaningfully in wording, sentence structure, tone, or emphasis.
- Suggested variants: (1) concise and clinical, (2) descriptive and narrative, (3) reorder the emphasis.
- No near-duplicates or trivial paraphrases.

Output format (MUST follow exactly):
- Return ONLY the three captions, nothing else.
- Each caption is on its own single line.
- Prefix each line with "1. ", "2. ", "3. " (number, period, space) — no other numbering style.
- No preamble, no labels like "Caption 1:", no bullets, no quotation marks, no blank lines between captions, no trailing commentary.

Example of the required output format (content is illustrative only):
1. First caption text here.
2. Second caption text here.
3. Third caption text here.
\end{quote}

Extended Table~\ref{tab:caption_diversity} reports corpus-level lexical diversity using template captions and LLM-generated captions, measured over all \(N=5{,}114\) captions from all 5 datasets without expert-annotated reports (including the 2 held-out datasets, but they are not involved in training). LLM rewriting substantially increased vocabulary breadth (unique unigrams: 302 $\to$ 729; unique bigrams: 964 $\to$ 4{,}525) and type-token ratio at both the unigram (\textsc{Distinct-1}: 0.0047 $\to$ 0.0074) and bigram level (\textsc{Distinct-2}: 0.0162 $\to$ 0.0486), while also producing longer, more descriptive captions (mean length: 12.7 $\to$ 19.2 words).

\begin{table}[h]
    \centering
    \caption{Lexical diversity statistics before and after LLM-based caption
             diversification ($N = 5{,}114$ captions).}
    \label{tab:caption_diversity}
    \begin{tabular}{lrr}
        \toprule
        \textbf{Metric} & \textbf{Template} & \textbf{LLM-generated} \\
        \midrule
        Total tokens          & 64{,}762  & 98{,}243  \\
        Unique unigrams       & 302       & 729       \\
        Unique bigrams        & 964       & 4{,}525   \\
        Distinct-1            & 0.0047    & 0.0074    \\
        Distinct-2            & 0.0162    & 0.0486    \\
        Mean length (words)   & 12.7      & 19.2      \\
        Median length (words) & 10        & 18        \\
        90th-pct.\ length     & 21        & 24        \\
        \bottomrule
    \end{tabular}
\end{table}

\subsection{Cross-modal Retrieval}
\label{subsec:retrieval}
\begin{table}[t]
\centering
\small
\setlength{\tabcolsep}{4pt}
\renewcommand{\arraystretch}{1.2}
\caption{Cross-modal retrieval performance on the held-out test set. Best results in each column are bolded.}
\label{tab:retrieval_results}
\begin{tabular}{lcccccc}
\toprule
& \multicolumn{3}{c}{\textbf{Image $\rightarrow$ Text (I2T)}} 
& \multicolumn{3}{c}{\textbf{Text $\rightarrow$ Image (T2I)}} \\
\cmidrule(lr){2-4} \cmidrule(lr){5-7}
\textbf{Model} & R@1 & R@5 & R@10 & R@1 & R@5 & R@10 \\
\midrule
\multicolumn{7}{l}{\textit{Benchmark models}} \\
OpenAI CLIP & 0.000 & 0.003 & 0.006 & 0.000 & 0.004 & 0.009 \\
BiomedCLIP & 0.005 & 0.016 & 0.024 & 0.006 & 0.016 & 0.032 \\
\midrule
\multicolumn{7}{l}{\textit{CLIP text encoder variants}} \\
MLP heads only & 0.024 & 0.080 & 0.130 & 0.021 & 0.084 & 0.134 \\
MLP + image encoder & 0.034 & 0.117 & \textbf{0.184} & 0.037 & 0.124 & 0.187 \\
MLP + text encoder & \textbf{0.036} & \textbf{0.120} & \textbf{0.184} & \textbf{0.037} & \textbf{0.126} & \textbf{0.191} \\
MLP + image + text encoder & 0.003 & 0.015 & 0.028 & 0.004 & 0.017 & 0.032 \\
\midrule
\multicolumn{7}{l}{\textit{BioClinicalBERT text encoder variants}} \\
MLP heads only (BERT) & 0.008 & 0.037 & 0.056 & 0.006 & 0.033 & 0.061 \\
MLP + image encoder (BERT) & 0.004 & 0.027 & 0.048 & 0.006 & 0.027 & 0.056 \\
MLP + text encoder (BERT) & 0.027 & 0.094 & 0.151 & 0.026 & 0.094 & 0.160 \\
MLP + image + text encoder (BERT) & 0.014 & 0.064 & 0.112 & 0.015 & 0.060 & 0.112 \\
\bottomrule
\end{tabular}
\end{table}

Extended Table~\ref{tab:retrieval_results} reports cross-modal retrieval performance on the held-out test set, evaluated using Recall@K (R@1, R@5, and R@10) for both image-to-text (I2T) and text-to-image (T2I) retrieval. These metrics measure the ability of the model to retrieve the correct paired modality within the top-K ranked candidates.

Overall, models based on the CLIP text encoder achieve stronger retrieval performance than those using BioClinicalBERT. In particular, the CLIP-based variant with a fine-tuned text encoder attains the best results across all Recall@K metrics for both I2T and T2I retrieval. This suggests that CLIP-style text representations are more effective for preserving global alignment structure in the joint embedding space, which is critical for ranking-based retrieval tasks.

It is important to note that these retrieval results are computed over the entire held-out test split, where a strict one-to-one correspondence between images and reports is not guaranteed for all samples. Although the majority of the dataset consists of expert-generated reports that are expected to align closely with their corresponding images, some degree of mismatch or ambiguity may still be present. This limitation can introduce noise into the retrieval evaluation and may partially explain the relatively low absolute Recall@K values observed across all models.

In future work, we plan to construct a curated subset of the dataset with explicitly verified one-to-one image–report pairs to provide a more reliable result for cross-modal retrieval.

\subsection{Extended Results: Classification Prompt Template Strategy}

\begin{figure}[t]
    \centering
    \includegraphics[width=\linewidth]{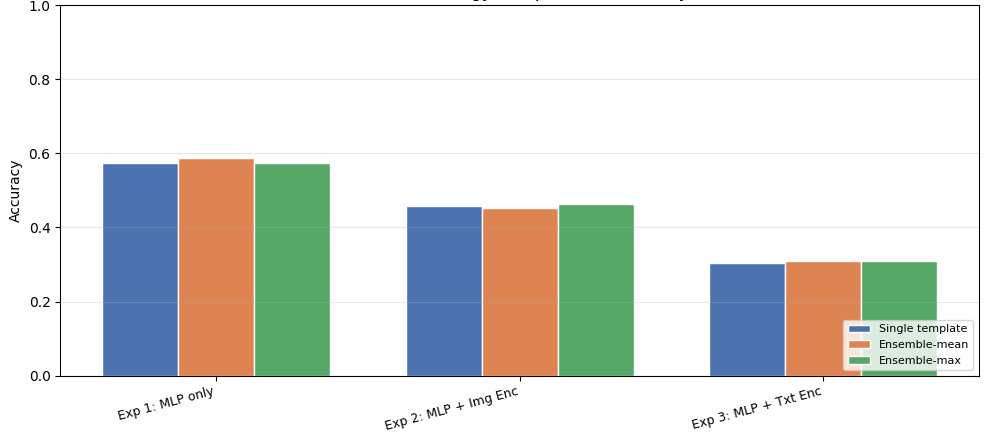}
    \caption{Zero-shot classification performance under three prompt template strategies: Single Template, Ensemble-Mean, and Ensemble-Max.}
    \label{fig:template_ablation}
\end{figure}

We conducted an ablation study to assess the impact of different prompt template strategies on zero-shot classification accuracy. Three strategies were evaluated: (1)~\textbf{Single Template}, in which a single fixed prompt is used for all classes; (2)~\textbf{Ensemble-Mean}, in which the text embedding for each class is computed as the mean of embeddings produced by ten distinct templates, each incorporating both tissue type and condition labels; and (3)~\textbf{Ensemble-Max}, in which the alignment score for each class is taken as the maximum across all ten per-class template embeddings. All strategies were evaluated under the same zero-shot classification protocol described in our previous Midterm Report. Results are shown in Extended Figure~\ref{fig:template_ablation}. Across all three strategies, performance differences were modest, suggesting that the model's zero-shot classification is largely robust to the choice of prompt template.

\end{document}